\relax
\documentclass[letterpaper]{article}
\usepackage{aaai18} 
\usepackage{times} 
\usepackage{helvet}  
\usepackage{courier}  
\usepackage{url}  
\usepackage{graphicx}  
\usepackage{amssymb}
\usepackage{amsmath}
\usepackage{amsfonts}
\usepackage{color}
\usepackage{algorithmic}
\usepackage{array}
\usepackage{subcaption}
\usepackage{float}
\usepackage[linesnumbered,ruled]{algorithm2e}
\usepackage[toc,acronym]{glossaries}
\usepackage{gensymb}
\usepackage{url}
\usepackage{amsthm}
\usepackage{gensymb}
\usepackage{ifthen,version}
\usepackage{tikz}
\usepackage{bm}

\begin{document}
\title{Adaptive Grasp Control through Multi-Modal Interactions for\\Assistive Prosthetic Devices}
\author{Michelle Esponda and Thomas M. Howard\\
Hajim School of Engineering and Applied Sciences\\
University of Rochester\\
Rochester, NY, 14627\\
}
\maketitle
\begin{abstract}
The hand is one of the most complex and important parts of the human body. The dexterity provided by its multiple degrees of freedom enables us to perform many of the tasks of daily living which involve grasping and manipulating objects of interest. Contemporary prosthetic devices for people with transradial amputations or wrist disarticulation vary in complexity, from passive prosthetics to complex devices that are body or electrically driven. One of the important challenges in developing smart prosthetic hands is to create devices which are able to mimic all activities that a person might perform and address the needs of a wide variety of users. The approach explored here is to develop algorithms that permit a device to adapt its behavior to the preferences of the operator through interactions with the wearer. This device uses multiple sensing modalities including muscle activity from a myoelectric armband, visual information from an on-board camera, tactile input through a touchscreen interface, and speech input from an embedded microphone. Presented within this paper are the design, software and controls of a platform used to evaluate this architecture as well as results from experiments deigned to quantify the performance. 
\end{abstract}

\section{Introduction}
Limb loss is a prevalent and growing issue in contemporary society. In the United States alone, there are currently two million people who live with limb loss. According to the Amputee Coalition, this is expected to double by 2050 \cite{Ziegler-Graham2008}. Dysvascular disease is the leading cause of lower limb amputations, while leading causes of upper limb amputations include trauma, cancer, and congenital disorders \cite{Ziegler-Graham2008}. Many people who have had amputations will consider using a prosthetic limb to improve their ability to perform many common tasks of daily living. Common options for upper limb amputees include passive, body-powered, and electrically powered prosthetics. Passive arm and hand prosthetics are made to be aesthetically pleasing but do not provide any joint movements for grasping objects. Body-powered arm and hand prosthetics are powered by the movement of the users arms or shoulders to drive cables that provide an open and close for grasping. This option is potentially more physically taxing on the user. Electrically powered arm and hand prosthetics are another option which utilize motors for finger movement. Many electrically powered prosthetics are controlled through a person's muscle signals. These are referred to as myoelectric prosthetics. 

\begin{figure}[htb]
	\begin{center}
	\includegraphics[width=0.47\textwidth]{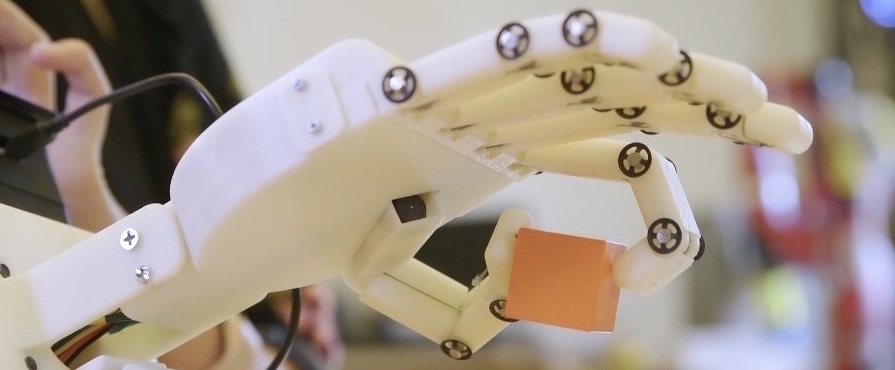}
	\caption{Multi-modal adaption of grasp control on a device exhibiting a pinch grasp on a small block.}
	\label{fig:grasping}
	\end{center}
\end{figure}

Myoelectric prosthetics offer a range of multi-articulated prosthetic hands. These utilize electrodes to read electromyographic (EMG) signals from the extensor and flexor muscles of the residual limb. Reading from EMG signals is difficult in part due to noise from electrical equipment, electrode movement, cross contamination from neighboring muscles, tissues, hair, and sweat. It is also difficult to read fine muscle signals, so myoelectric prosthetics tend to use muscle activities that are easy to discern, such as a general flex and extend pattern to classify different grasp actions. This requires the user to learn a mapping of muscle contraction patterns to desired hand activities. Reading EMG signals from a person with muscle dystrophy increases the difficulty of utilizing myoelectric signals since their muscle signals tend to be very faint, which makes it difficult to observe and classify. 

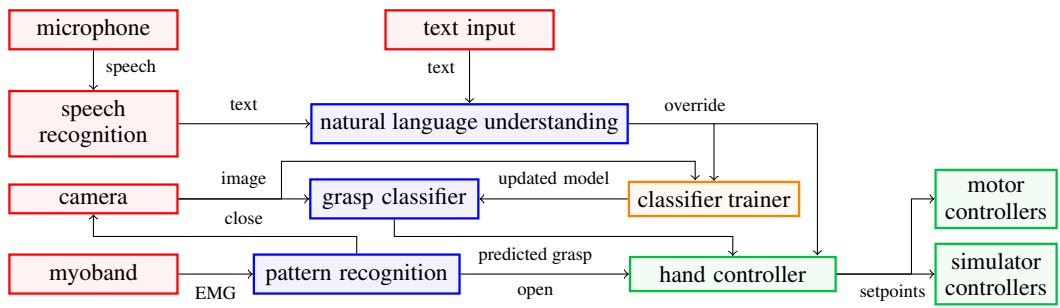
\begin{figure*}[htb]
\begin{center}
\begin{tikzpicture}
\node[rectangle,draw=red, fill=red!5, line width=0.03cm, text width=2cm,font=\footnotesize,align=center] at (0,3.25) (microphone) {microphone};
\node[rectangle,draw=red, fill=red!5, line width=0.03cm, text width=2cm,font=\footnotesize,align=center] at (0,2.0) (speechrecognition) {speech recognition};
\node[rectangle,draw=red, fill=red!5, line width=0.03cm, text width=2cm,font=\footnotesize,align=center] at (5,3.25) (textinput) {text input};
\node[rectangle, draw=red, fill=red!5, line width=0.03cm, text width=2cm,font=\footnotesize,align=center] at (0,1.0) (camera) {camera};
\node[rectangle, draw=red, fill=red!5, line width=0.03cm, text width=2cm,font=\footnotesize,align=center] at (0,0.0) (myoband) {myoband};
\node[rectangle, draw=blue, fill=blue!5, line width=0.03cm, text width=2cm,font=\footnotesize,align=center] at (4,1.0) (graspclassifier) {grasp classifier};
\node[rectangle, draw=blue, fill=blue!5, line width=0.03cm, font=\footnotesize,align=center] at (5,2.0) (naturallanguageunderstanding) {natural language understanding};
\node[rectangle,draw=blue, fill=blue!5, line width=0.03cm, text width=2.5cm,font=\footnotesize,align=center] at (3.5,0.0) (patternrecognition) {pattern recognition};
\node[rectangle,draw=green!75!blue, fill=green!75!blue!5, line width=0.03cm, text width=2.5cm,font=\footnotesize,align=center] at (8.5,0.0) (handcontroller) {hand controller};
\node[rectangle,draw=green!75!blue, fill=green!75!blue!5, line width=0.03cm, text width=1.375cm,font=\footnotesize,align=center] at (12,1.0) (motorcontrollers) {motor controllers};
\node[rectangle,draw=green!75!blue, fill=green!75!blue!5, line width=0.03cm, text width=1.375cm,font=\footnotesize,align=center] at (12,0.0) (simcontrollers) {simulator controllers};
\node[rectangle,draw=orange, fill=orange!5, line width=0.03cm, font=\footnotesize,align=center] at (8.25,1.0) (classifiertrainer) {classifier trainer};
\draw[->] (microphone) to (speechrecognition);
\draw[->] (speechrecognition) to (naturallanguageunderstanding);
\draw[->] (textinput) to (naturallanguageunderstanding);
\draw[->] (camera) to (graspclassifier);
\draw[->] (myoband) to (patternrecognition);
\draw[->] (8.25,2) to (9.625,2) to (9.625,0.25);
\draw[->] (graspclassifier) to (4,0.5) to (8.5,0.5) to (handcontroller);
\draw[->] (patternrecognition) to (handcontroller);
\draw[->] (handcontroller) to (10.875,0) to (10.875,1) to (motorcontrollers);
\draw[->] (handcontroller) to (10.875,0) to (simcontrollers);
\draw[->] (classifiertrainer) to (graspclassifier);
\draw[->] (naturallanguageunderstanding) to (8.25,2) to (classifiertrainer);
\draw[->] (camera) to (2.5,1) to (2.5,1.5) to (8,1.5) to (8,1.25);
\draw[->] (patternrecognition) to (3.5,0.5) to (0,0.5) to (camera);
\node[font=\scriptsize,align=center] at (0.5,2.75) (speech) {speech};
\node[font=\scriptsize,align=center] at (2,2.25) (text1) {text};
\node[font=\scriptsize,align=center] at (4.625,2.75) (text2) {text};
\node[font=\scriptsize,align=center] at (2,1.25) (image) {image};
\node[font=\scriptsize,align=center] at (1.625,-0.25) (emg) {EMG};
\node[font=\scriptsize,align=center] at (5.875,-0.25) (open) {open};
\node[font=\scriptsize,align=center] at (2,0.75 )(close){close};
\node[font=\scriptsize,align=center] at (10.625,-0.25) (setpoints) {setpoints};
\node[font=\scriptsize,align=center] at (5.875, 0.25) (predictedgrasp) {predicted grasp};
\node[font=\scriptsize,align=center] at (8,2.25) (override) {override};
\node[font=\scriptsize,text width=2.5cm,align=center] at (6.125,1.25) (updatedmodel) {updated model};
\end{tikzpicture}
\end{center}
\caption{The proposed system architecture for an adaptive multi-modal prosthetic hand.}
\label{fig:architecture}
\end{figure*}

Recent advances in sensors and algorithms in robotics provide a number of opportunities for enhancing the performance of myoelectric-based prostheses, including the ability to sense the environment to provide context to the interpretation of muscle signals. Recent advances in language understanding also enable humans to provide natural corrections or feedback to the system to adjust the characteristics of the controller. This provides an opportunity to incorporate visual and audial information into the controller that can adapt to the preferences of the user to create a more robust, responsive and potentially more reliable prosthetic hand.  

In this paper we present a novel system architecture for multi-modal control and adaptation of a prosthetic hand. A visual representation of this architecture is illustrated in Figure \ref{fig:architecture}. EMG, visual, and audial data is interpreted, classified, and utilized by a grasping controller. The behavior of the controller is governed by an artificial neural network that predicts optimal grasp shapes to be executed upon observations of flexion or extension of the forearm muscles read by the EMG sensor. Corrections to this model are provided from tactile input via a forearm-mounted touchscreen or through language via a wrist-mounted microphone. Newly acquired training examples feed the predictive model of preferred grasp shape which is incorporated into the model for future manipulations of these objects.  

\section{Background}
There exists a multitude of ways to control prosthetic devices, including using muscle signals, computer vision, or audio. Due to the interest in myoelectrically controlled prosthetic hands, how to best extract features and classify the EMG signals has been studied. Shenoy, et al, utilizes RMS for feature extraction and SVM for classification for controlling a robotic arm \cite{Shenoy2008}. Others, such as Chan and Englehart explore Hidden Markov Models to achieve higher classification for control of upper limb prostheses \cite{Chan2005}. Ju, et al, presents Fuzzy Gaussian Mixture Models as an approach for nonlinear classification of prosthetic hand grasps \cite{Ju2013}. Many other approaches attempt to create a hybrid model with EMG signals and other sensors. Atzori, et al, presents the benefit of using additional sensors for control by adding accelerometers to their hand control model \cite{Atzori2014}. 

Vision-based modes of prosthetic hand control have also been explored. With the use of a camera, an object can be detected and classified to a specific grip pattern. The IRIS hand is an example of one such project \cite{Casley2014}. This prosthetic was developed with an integrated camera at the palm. The IRIS hand segments the object from its background using Canny Edge detection algorithm and identifies and labels the object using Hough Transformation and SURF detection. Another approach uses grayscale images which are preprocessed using median and Gaussian filters for smoothing, segmented using automatic thresholding, and target selection by selecting the center most object of the image \cite{Dosen2010}. The CamHand uses Convolutional Neural Networks (CNN) when classifying an object to a grasp \cite{DeGol2016}. A hybrid control approach uses augmented reality (AR) glasses with stereo cameras triggered by myoelectric control \cite{Marko2014}. The image being viewed is processed and the selected grasp and aperture size for grasping is shown to the user through the glasses. The user has the option to correct this selection through myoelectric control.  

Computer vision based control models are not only used to classify objects to set grasp types, but are also utilized for grasp planning and manipulation. Automated grasping is used to plan out finger movements and positions on an object. Bender and Bone create an automated grasp planing process which obtains a 2D model of the object using computer vision and inputs this into the grasp planner \cite{Bender2004}. The robotic hand probes the object for tactile information, such as height, to allow for 2.5D grasping. The shape, position, and orientation of the objects are originally unknown. Saxena, et al, provides another approach for grasping novel objects. This research chose to label grasping points on household objects using two or more 2D images of the object at different perspectives to obtain, not the 3D model of the object, but the 3D grasping points \cite{Saxena2008}. GraspIt \cite{Kragic2001} is a grasp planning and visualization system developed to automate grasping of general objects. Kragic, et al uses the objects pose using a CAD model of the object and plans the grasp technique and generates trajectories to move the fingers.

The last approach for prosthesis hand control discussed here is through language and voice recognition. Mainardi and Davalli explore this in their research where they control a prosthesis using a throat microphone \cite{Mainardi2007} to reduce noise in the room as well as to perform complex tasks more quickly than with a myoelectically controlled prosthetic. Gruppioni E., et al, discusses voice control and the appropriate vocabulary to be used for controlling the shoulder, elbow, wrist, and hand of a prosthesis \cite{Gruppioni2008}. This research takes note in selecting words which are distinct enough for the voice controller to recognize. Other research in prosthetic language control include a Digital Signal Processor- based system that takes in spoken commands to control an artificial limb \cite{Lin1998}.  Although language based control of prosthetic hands continues to be a novel and minimally explored research area, language based control in robotics continues to be a large and growing field of research. Much of the research has focused on the development of efficient algorithms for unidirectional or bi-directional human-robot interaction through natural language interactions \cite{tellex11a,boteanu2017robotinitiatedinteraction,paul2018efficientplatforms}.  Broad, et al, investigates correcting the behavior and constraints of a robotic arm manipulator to perform certain tasks using speech-to-text software and a Distributed Correspondence Graph (DCG) for natural language understanding \cite{broad17a}. The novel aspects of the multi-modal interface for adaptive grasp control are most closely related to the corrections provided through natural language interaction in \cite{broad17a}, however corrections available through both touchscreen and language interfaces provide examples for adaptation of the grasping controller behavior over time instead of only updating the current constraints. 

\section{Experimental Platform}
To test our approach to language-guided adaptation of grasping control for assistive prosthetic devices, we designed and constructed a 3D-printed device that mimics the anatomy of a five-fingered hand/wrist. This contains sensors to provide observations, computing to classify observations and infer grasp controllers, and actuators to drive physical interactions with objects in the environment.  This section will discuss the electromechanical design for the device created to evaluate the proposed system architecture.  

\begin{figure}[htb]
\begin{center}
  \includegraphics[width=0.47\textwidth]{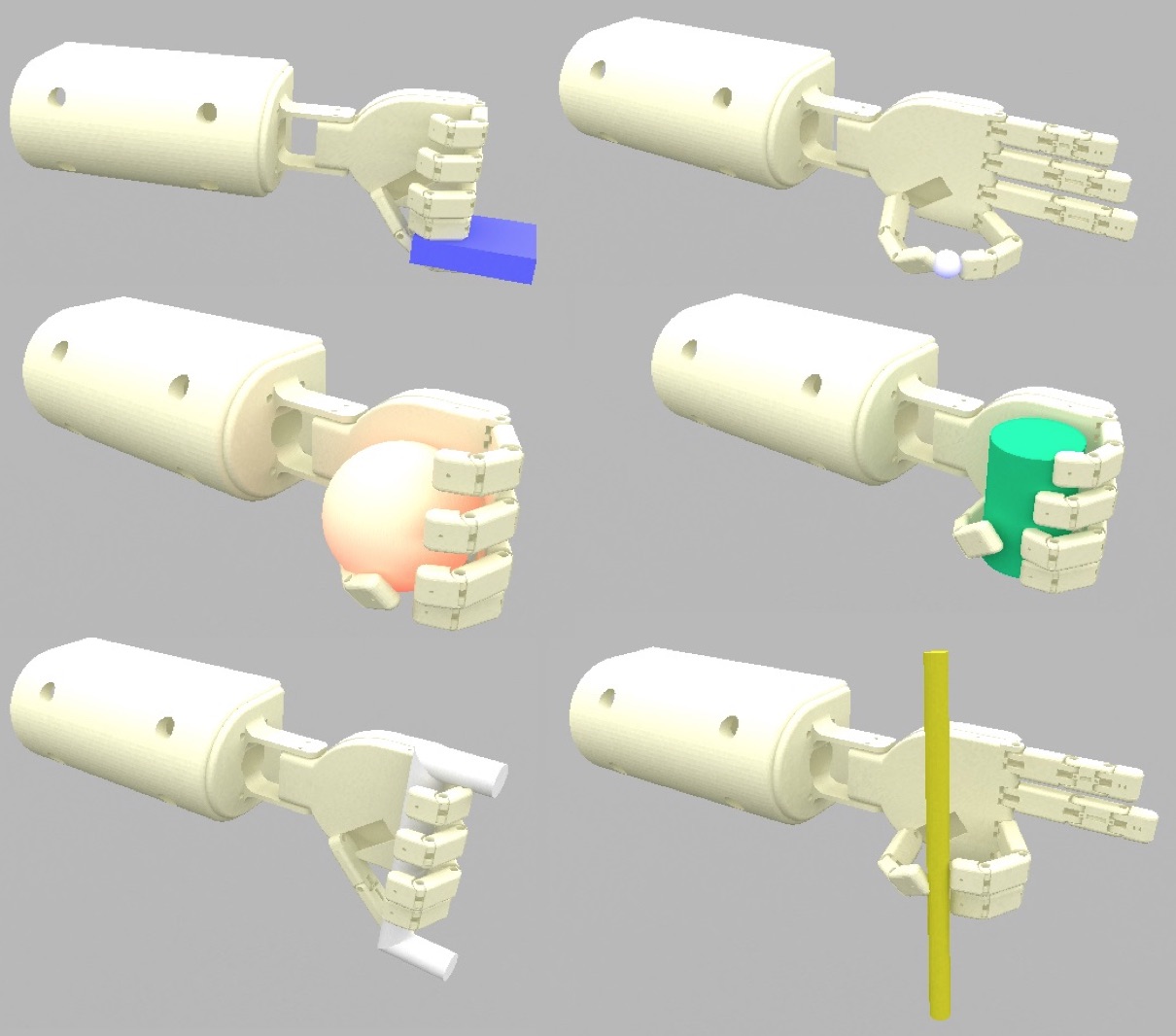}
	\begin{picture}(0,0)
	\put(-100,170){\scriptsize{lateral grasp}}
	\put(-100,105){\scriptsize{spherical grasp}}
	\put(-100,40){\scriptsize{hook grasp}}
	\put(0,170){\scriptsize{pinch grasp}}
	\put(0,105){\scriptsize{cylindrical grasp}}
	\put(0,40){\scriptsize{tripod grasp}}
	\end{picture}
	\caption[Simulation of the prosthetic hand performing various grasp types]{Simulation of the prosthetic hand performing various grasp types. The simulated hand can be seen here performing a cylindrical, hook, tripod, spherical, lateral, and pinch grasp.}
	\label{fig:simulation}
	\end{center}
\end{figure}
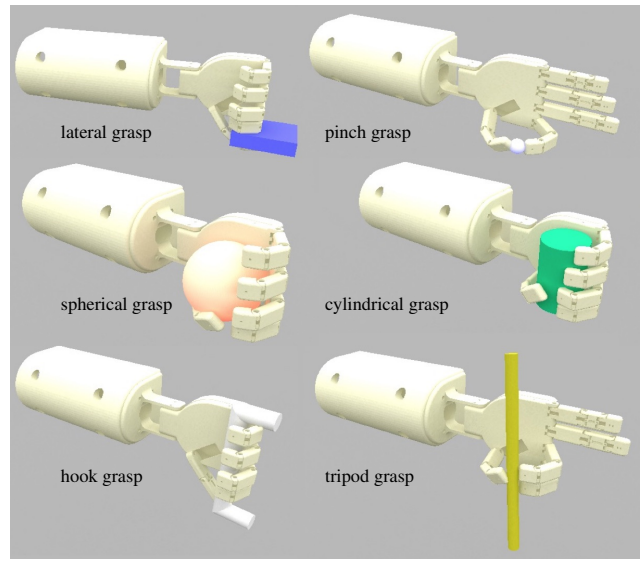

A number of researchers have studied the problem of grasping for both humans and robots \cite{Yang2015,Cutkosky1989,Feix2015,Iberall1997}. According to Thomas Feix, et al, the most common grasp type tends to be a medium wrap, followed by a lateral pinch. The third most common grasp was found to be a thumb plus two finger grasp \cite{Feix2014}. These are categorized more broadly into power and precision grasps. Power grasps are used for heavy, rigid objects which require more force to hold. Precision grasps are for a more precise and accurate handling of an object. The six most common hand grasps we designed the hand to be capable of executing include cylindrical, spherical and hook power grasps and lateral, pinch, and tripod precision grasps. The dimensions and mounting location of the fingers were designed to ensure that the prosthetic would be capable of executing all of these grasps for a set of common objects. To test this during the design phase, iterations of the design were simulated in the context of various objects placed in configurations near the palm. Figure \ref{fig:simulation} illustrates a simulation of the final design grasping with each of the six different grasp types for different objects.

There exists a number of robot hands that are suitable for electrically controlled prosthetic devices, including open-source projects based on innovative cable-driven designs described in \cite{ma2013a}.  The design of the hand we developed to test our architecture for adaptive grasp control consists of distal, intermediate, and proximal phalanx for each of the four fingers, and a distal, proximal, and metacarpal phalanx for the thumb where the dimensions were guided by the principles of \cite{Alexander2010}. The flexion and extension of the fingers were enabled by a braided line cord threaded through a canal designed into the anterior portion of the fingers. These cords are drawn by a motor to close the fingers and represent the behavior of the flexor tendons of the hand. To extend the fingers, 0.42 in-lb torsion springs are placed at the joints. The final design of the thumb allows forward abduction, opposition, re-position, and flexion and extension. The design includes a wrist mount for testing and an enclosure-mounted screen containing a graphical user interface (GUI), described in a later section for data collection, is fastened on top.

Six 130 RPM micro gear motors are used to actuate the prosthetic hand. The first five motors control the shape of each finger by pulling on the tendon cords while the last motor controls the direction of the thumb. The incremental encoders on each of the motors are quadrature encoders that use magnetic hall effect sensors, giving feedback on the direction and position of all motors. L293D dual H-bridge drivers are also utilized, controlling the direction of the gear motors.  Speed control is achieved by setting the duty cycle for each motor using pulse-width modulation (PWM). A PID controller sets the duty cycle using the observed error between the desired and measured angle. Power to the motors is provided by a 1800 mAh Ni-MH battery pack.

The Raspberry Pi 3 Model B micro computer using Raspbian OS is used as the principal computer inside of the adaptive prosthetic hand prototype. The forty GPIO pins of the Raspberry Pi 3 are utilized to power the drivers, encoders, and touchscreen, control the motors and drivers, and read from the encoders.  For sensing, we used the PiCam, a USB-based microphone and a Myo Armband through the Bluetooth interface.  To display the graphical user interface and interact with the software, we used a seven inch touchscreen interface.  Interprocess communication between software modules is performed using LCM \cite{huang10a}.  Power to the computer is provided by a 4400 mAh battery pack.

\begin{figure}[htb]
\begin{center}
	\mbox{
	\begin{subfigure}[b]{0.42\textwidth}
		\includegraphics[width=\textwidth]{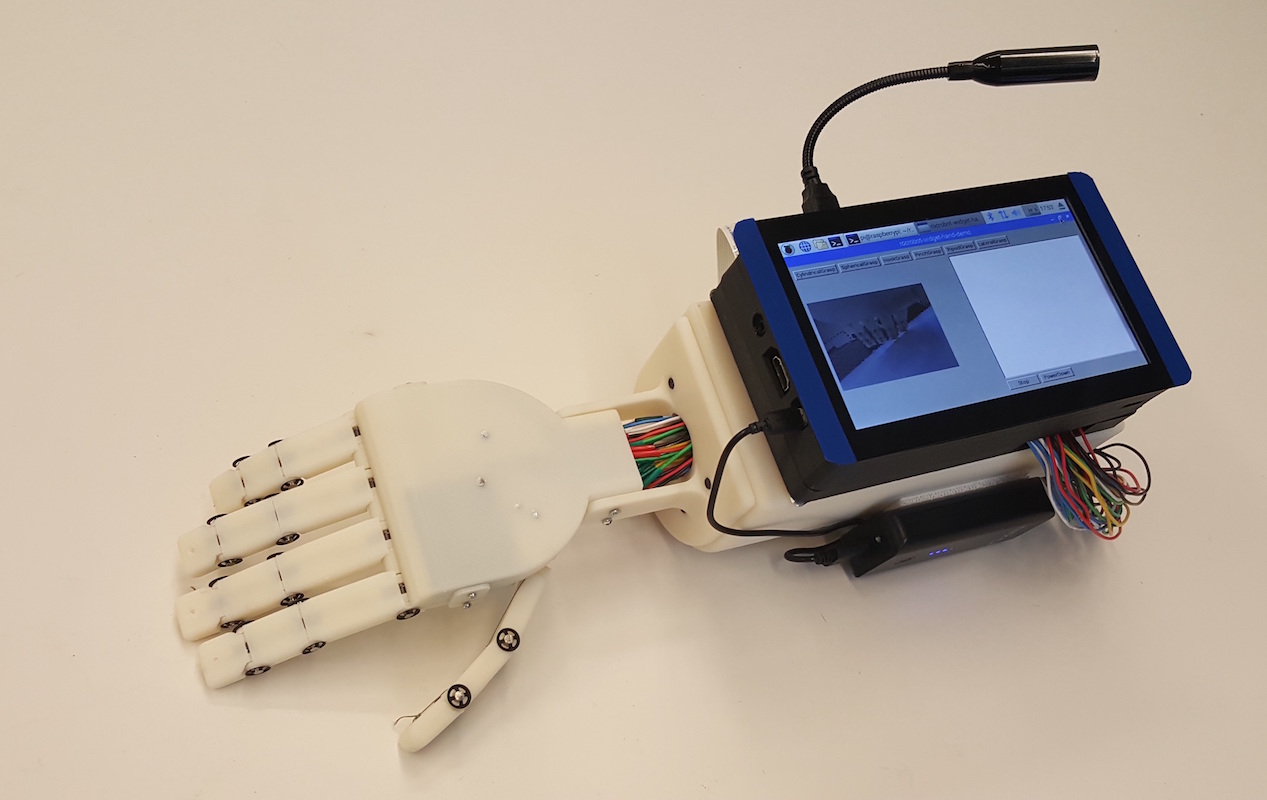}
		\caption{Adaptive prosthetic hand showing the USB microphone}
	\end{subfigure} 
	} \\[8pt]
	\mbox{
	\begin{subfigure}[b]{0.2\textwidth}
		\includegraphics[width=\textwidth]{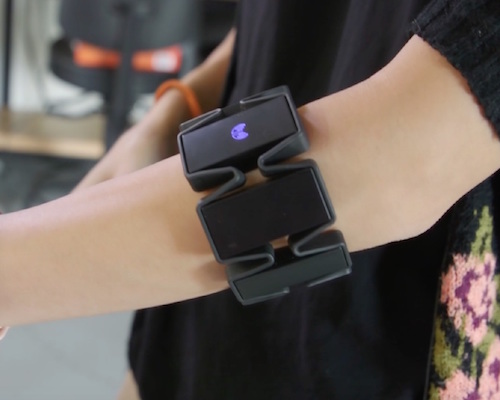}
		\caption{sEMG armband}
	\end{subfigure} \,
	\begin{subfigure}[b]{0.2\textwidth}
		\includegraphics[width=\textwidth]{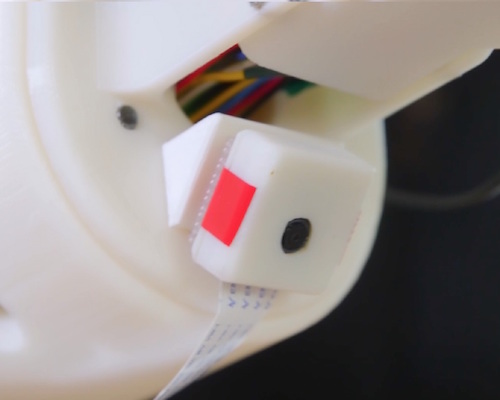}
		\caption{Pi Camera}
	\end{subfigure}
	}
	\caption{Full system architecture with USB microphone, the sEMG armband worn at the appropriate position on the forearm, and Pi camera mounted at the base.}
	\label{fig:sensors}
\end{center}
\end{figure}

\section{Sensor Classification and Model Adaptation}
The architecture for adaptive grasping control from multi-modal inputs must be capable of interpreting sEMG, visual, and audial information to inform grasp behaviors and receive and utilize new examples in-situ.  This section describes the applied classifiers and the input for the operator.

EMG classification is performed by a Myo Armband produced by Thalmic Labs, which contains eight electrodes to read from the flexor and extensor muscles, and is worn on the forearm just below the elbow as shown in Figure \ref{fig:sensors}. Five gestures are classified by the armband: fist, spread fingers, wave in, wave out, and finger tap. Calibration is performed by performing a specific hand gesture each time the armband is placed on by the operator. The classification done by the Myo Armband is utilized for this project. The Myo Armband pose process from the software architecture described in Figure \ref{fig:sensors} publishes either an open hand (spread fingers) or close hand (fist) message on a pose channel. The open and close hand signals are used to initiate the prosthetic hand to either open its fingers or to decide on a grasp type. With the close hand signal, the neural network process classifies an incoming image from the camera to a particular grasp type. This grasp type is published to the motor controls to move the fingers to the appropriate positions.

Grasp classification is performed by predicting the most probable grasp from downsampled images acquired from the 8-megapixel Raspberry Pi Camera Module V2, which is mounted at the base of the prosthetic hand as pictured in Figure \ref{fig:sensors}. A feed-forward artificial neural network is used to classify the image to a grasp type. The input layer consists of a vectorized 80 x 60 RGB color image. The two hidden layers of the neural network contain 300 and 50 nodes respectively. The input features go through the various layers and the proper weights produced. Weights are trained in MATLAB using Scaled Conjugate Gradient \cite{Moller1990}. Fifty percent of the images are used for training, while a separate twenty-five percent used for validation and twenty-five percent for testing. The output layer consists of six values that represent the estimated posterior probability of each particular grasp. The system is ran at five folds, with the best outcome picked and its weights saved to a data file. The grasp classifier in Figure \ref{fig:architecture} emits a message with the most likely grasp to the grasp controller to execute.

To collect and label images to a specific grasp type for the model, an interactive Qt-based GUI was created.  The GUI contains a live feed of the downsampled camera image, six buttons representing the different hand grasps used in this project (cylindrical, spherical, hook, pinch, tripod, and lateral grasps), and a read-only text box for feedback. Pressing a grasp button publishes messages to save the image of the object to the desired grasp for retraining the grasp classification model and execute the desired grasp. The touchscreen is used to view the GUI and is placed over the wrist mount of the prosthetic hand, shown in Figure \ref{fig:sensors}. 

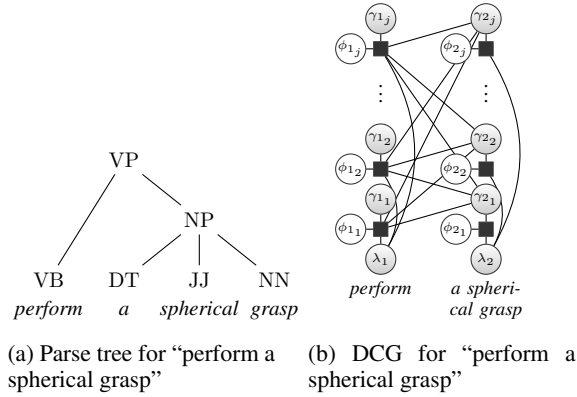
\begin{figure}[htb]
 \begin{center}
\begin{subfigure}[b]{0.20\textwidth}
\centering
\begin{tikzpicture}[scale=0.80, transform shape]
\node[] (vb1) at (0,1.0) {$\mathrm{VB}$};   
\node[] (dt1) at (1.25,1.0) {$\mathrm{DT}$};     
\node[] (jj1) at (2.5,1.0) {$\mathrm{JJ}$};
\node[] (nn1) at (3.75,1.0) {$\mathrm{NN}$};    
\node[] (np1) at (2.5,2.0) {$\mathrm{NP}$};
\node[] (vp1) at (1.25,3.0) {$\mathrm{VP}$};
\node[anchor=mid](l1) at (0,0.5) {\textit{perform}};
\node[anchor=mid](l2) at (1.25,0.5) {\textit{a}};
\node[anchor=mid](l3) at (2.5,0.5) {\textit{spherical}};
\node[anchor=mid](l4) at (3.75,0.5) {\textit{grasp}};
\draw[-] (dt1) -- (np1);  
\draw[-] (jj1) -- (np1);      
\draw[-] (nn1) -- (np1);     
\draw[-] (vb1) -- (vp1);
\draw[-] (np1) -- (vp1);
\end{tikzpicture}
\caption{Parse tree for ``perform a spherical grasp''}
\label{fig:nlu-parse}
\end{subfigure} \quad
\begin{subfigure}[b]{0.20\textwidth}
\centering
\begin{tikzpicture}[scale=0.80, transform shape, textnode/.style={anchor=mid,font=\small},
      nodeknown/.style={circle,draw=black!80,fill=black!10,minimum size=5mm,font=\tiny,top color=white,bottom color=black!20},
      nodeunknown/.style={circle,draw=black!80,fill=white,minimum size=5mm,font=\tiny},
      factor/.style={rectangle,draw=black!80,fill=black!80,minimum size=2mm,font=\tiny,text=white},
      dot/.style={circle,draw=black!80,fill=black!80,minimum size=0.25mm,font=\tiny}]
    \pgfdeclarelayer{background}
    \pgfdeclarelayer{foreground}
    \pgfsetlayers{background,main,foreground}
    %
    %
    \node[textnode] (l1) at (0.0,0.0) {\textit{perform}};
    \node[textnode,text width=2cm, align=center] (l2) at (1.75,0.0) {\textit{a spherical grasp}};
  \draw (0.0,3.125) node {.};
  \draw (0.0,3.25) node {.};
  \draw (0.0,3.375) node {.};
  \draw (1.75,3.125) node {.};
  \draw (1.75,3.25) node {.};
  \draw (1.75,3.375) node {.};
    \node[nodeknown] (vp1) at (0.0,0.5) {};
    \node[nodeknown] (np1) at (1.75,0.5) {};
    \node[font=\tiny] (l1label) at (0.0,0.5) {$\lambda_{1}$};
    \node[font=\tiny] (l2label) at (1.75,0.5) {$\lambda_{2}$};
    \node[nodeunknown] (c11) at (-0.5,1) {};
    \node[nodeunknown] (c12) at (-0.5,2) {};
    \node[nodeunknown] (c1n) at (-0.5,4) {};
    \node[nodeunknown] (c21) at (1.25,1) {};
    \node[nodeunknown] (c22) at (1.25,2) {};
    \node[nodeunknown] (c2n) at (1.25,4) {};
    \node[font=\tiny] (c11label) at (-0.5,1) {$\phi_{1_{1}}$};
    \node[font=\tiny] (c12label) at (-0.5,2) {$\phi_{1_{2}}$};
    \node[font=\tiny] (c1nlabel) at (-0.5,4) {$\phi_{1_{j}}$};
    \node[font=\tiny] (c21label) at (1.25,1) {$\phi_{2_{1}}$};
    \node[font=\tiny] (c22label) at (1.25,2) {$\phi_{2_{2}}$};
    \node[font=\tiny] (c2nlabel) at (1.25,4) {$\phi_{2_{j}}$};
    \node[nodeknown] (g11) at (0.0,1.5) {};
    \node[nodeknown] (g12) at (0.0,2.5) {};
    \node[nodeknown] (g1n) at (0.0,4.5) {};
    \node[nodeknown] (g21) at (1.75,1.5) {};
    \node[nodeknown] (g22) at (1.75,2.5) {};
    \node[nodeknown] (g2n) at (1.75,4.5) {};
    \node[font=\tiny] (g11label) at (0.0,1.5) {$\gamma_{1_{1}}$};
    \node[font=\tiny] (g12label) at (0.0,2.5) {$\gamma_{1_{2}}$};
    \node[font=\tiny] (g1nlabel) at (0.0,4.5) {$\gamma_{1_{j}}$};
    \node[font=\tiny] (g21label) at (1.75,1.5) {$\gamma_{2_{1}}$};
    \node[font=\tiny] (g22label) at (1.75,2.5) {$\gamma_{2_{2}}$};
    \node[font=\tiny] (g2nlabel) at (1.75,4.5) {$\gamma_{2_{j}}$};
    \node[factor] (f11) at (0.0,1) {};
    \node[factor] (f12) at (0.0,2) {};
    \node[factor] (f1n) at (0.0,4) {};
    \node[factor] (f21) at (1.75,1) {};
    \node[factor] (f22) at (1.75,2) {};
    \node[factor] (f2n) at (1.75,4) {};
    \begin{pgfonlayer}{background}
      \draw[-] (vp1) to (f11);
      \draw[-] (np1) to (f21);
      \draw[-] (vp1) to [bend right=30] (f12);
      \draw[-] (np1) to [bend right=30] (f22);
      \draw[-] (vp1) to [bend right=30] (f1n);
      \draw[-] (np1) to [bend right=30] (f2n);
      \draw[-] (c11) to (f11);
      \draw[-] (c21) to (f21);
      \draw[-] (c12) to (f12);
      \draw[-] (c22) to (f22);
      \draw[-] (c1n) to (f1n);
      \draw[-] (c2n) to (f2n);
      \draw[-] (g11) to (f11);
      \draw[-] (g21) to (f21);
      \draw[-] (g12) to (f12);
      \draw[-] (g22) to (f22);
      \draw[-] (g1n) to (f1n);
      \draw[-] (g2n) to (f2n);
      \draw[-] (g21) to (f11);
      \draw[-] (g21) to (f12);
      \draw[-] (g21) to (f1n);
      \draw[-] (g22) to (f11);
      \draw[-] (g22) to (f12);
      \draw[-] (g22) to (f1n);
      \draw[-] (g2n) to (f11);
      \draw[-] (g2n) to (f12);
      \draw[-] (g2n) to (f1n);
    \end{pgfonlayer}
  \end{tikzpicture}
\caption{DCG for ``perform a spherical grasp''}
\label{fig:nlu-dcg}
\end{subfigure}
\end{center}
\caption{The parse tree and DCG model for the utterance ``perform a spherical grasp''.}
\label{fig:nlu}
\end{figure}

To enable the operator to control the hand without using the GUI, desired grasping behaviors are inferred by a natural language understanding (NLU) process that uses text from speech acquired by a USB microphone as shown in Figure \ref{fig:sensors}.  Recognized speech is observed by listening for a keyword and processing the following output.  The classified text is parsed into a tree using a CYK parser \cite{Younger1967} as shown in Figure \ref{fig:nlu-parse} for the verbal command ``perform a spherical grasp''. 

Once the sentence is parsed, a probabilistic graphical model is formed from this structure to infer the most likely symbols in an application-dependent symbolic representation. Figure \ref{fig:nlu-dcg} illustrates the resulting Distributed Correspondence Graph (DCG) from the parse tree, a model developed for natural language understanding of robot instructions \cite{howard14a}.  The symbols $\lambda_{1}$ ... $\lambda_{n}$ represent each of the phrases while $\gamma_{1}$ ... $\gamma_{n}$ represent the groundings for these phrases (one of the six grasp types). The number of phrases is represented by \textit{N} and set of groundings by $\Gamma$. The binary correspondence variables are represented as $\phi_{1}$ ... $\phi_{n}$. Equation \ref{eq:dcgeqn} illustrates the search for the most probable correspondence variables given the phrases, groundings, child correspondence variables, and the environment ($\Upsilon$).  
\begin{equation}
\Phi^{*} = \operatorname*{arg\,max}_{\phi_{ij} \in \Phi} \prod_{i=1}^{|N|} \prod_{j=1}^{|\Gamma|} p(\phi_{ij} | \gamma_{ij}, \lambda_{ij}, \Phi_{c_{i}}, \Upsilon)
\label{eq:dcgeqn}
\end{equation}

The value of ``i'' represents the amount of phrases, while ``j'' represents the amount of possible groundings, in this case six. The model is trained from a corpus of annotated instructions that map natural language to grasping actions, such as ``a three fingered grasp'' mapping to a tripod grasp. The resulting ``true'' correspondence variables at the root of the sentence indicate activation of the grasp corresponding to the equivalent grounding. This grasp is then published to the grasp controller as well as to the camera to save the image of the object to the corresponding grasp type.   

\section{Experimental Design}
To verify and quantify the performance of the adaptive prosthetic hand, two experiments were conducted. The first experiment evaluates the function of the multi-modal inputs (GUI and NLU) for collecting and labeling new data and controlling the gripper. The second experiment evaluates the ability of the integrated system to adapt the model to new data provided by one of the input methods. 

The first experiment evaluates the ability of the adaptive prosthetic hand to accept inputs from touchscreen and audial inputs available to the operator. For both inputs, the prosthetic hand is guided to a position until the object is in view of the camera by utilizing the live feed of the camera shown on the GUI. For the touchscreen inputs, the preferred grasp type is selected by pressing the appropriate widget button on the GUI. For the audial input, audio signals are converted to speech and preferred grasp types are inferred from the NLU.  Phrases and sentences that did and did not appear in the training corpus are considered to evaluate how the system would respond to novel inputs. For both the touchscreen and audial inputs images are acquired corresponding to the particular grasp and the hand is expected to execute the specified grasp. This is repeated for a fixed number of times for each object and the success/failure of the grasp and data collection and the runtime of the NLU is measured.   

\begin{figure}[htb]
\begin{center}
\mbox{
    \begin{subfigure}[b]{0.11\textwidth}
	\includegraphics[width=\textwidth]{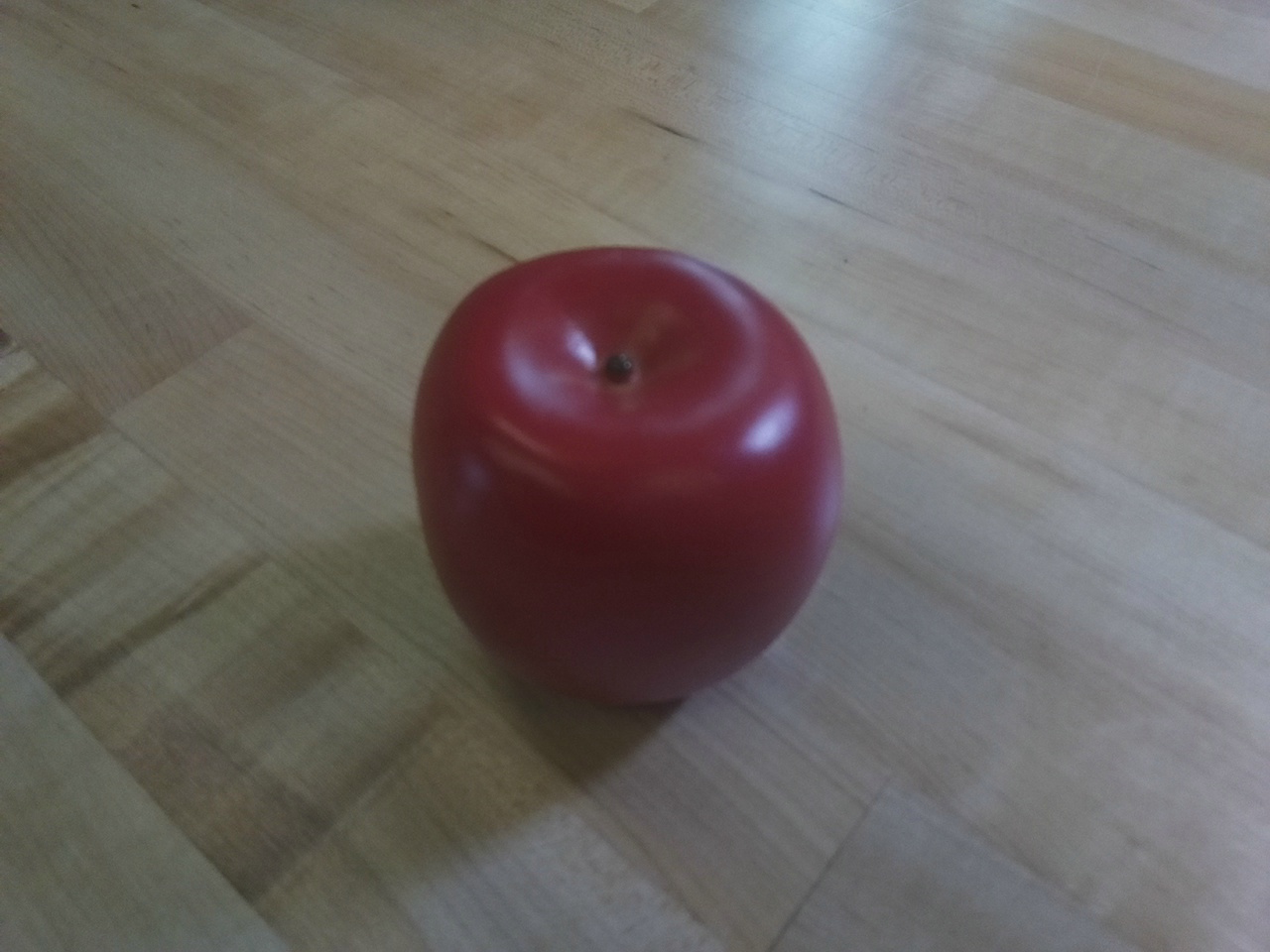}
	\caption{apple}
	\end{subfigure}
	\begin{subfigure}[b]{0.11\textwidth}
	\includegraphics[width=\textwidth]{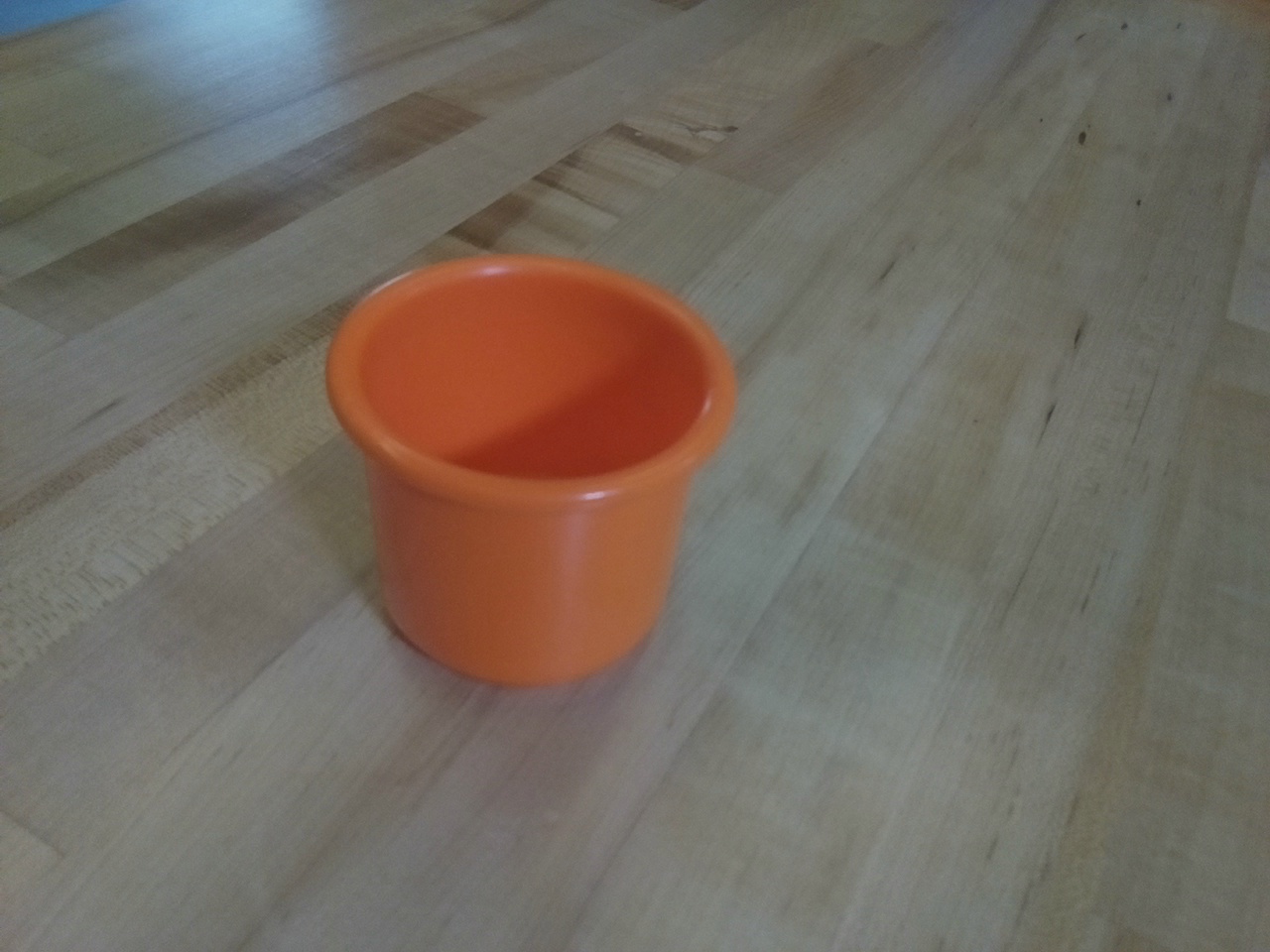}
	\caption{cup}
	\end{subfigure}
	\begin{subfigure}[b]{0.11\textwidth}
	\includegraphics[width=\textwidth]{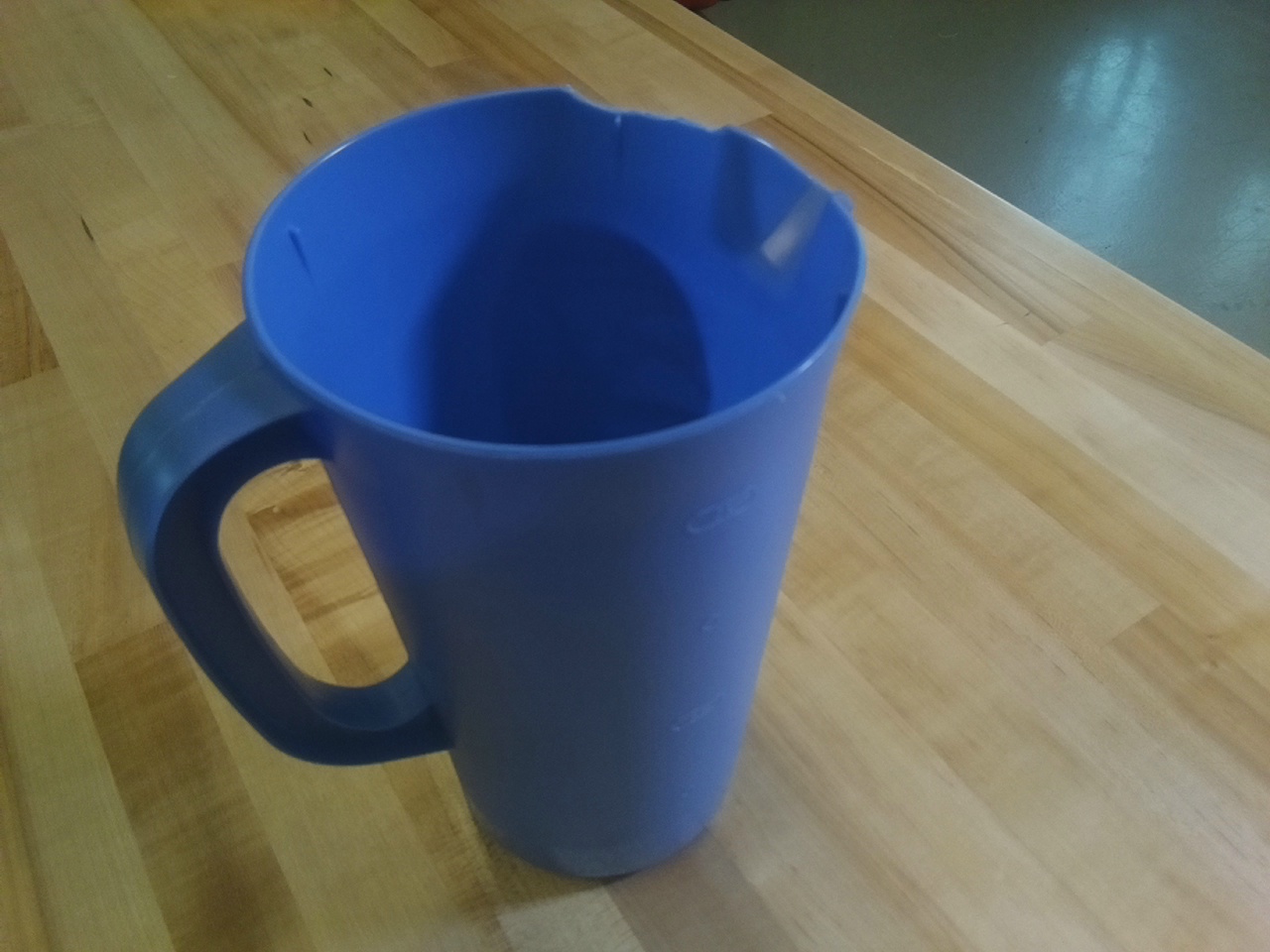}
	\caption{pitcher}
	\end{subfigure}
} \\[8pt]
\mbox{
    \begin{subfigure}[b]{0.11\textwidth}
	\includegraphics[width=\textwidth]{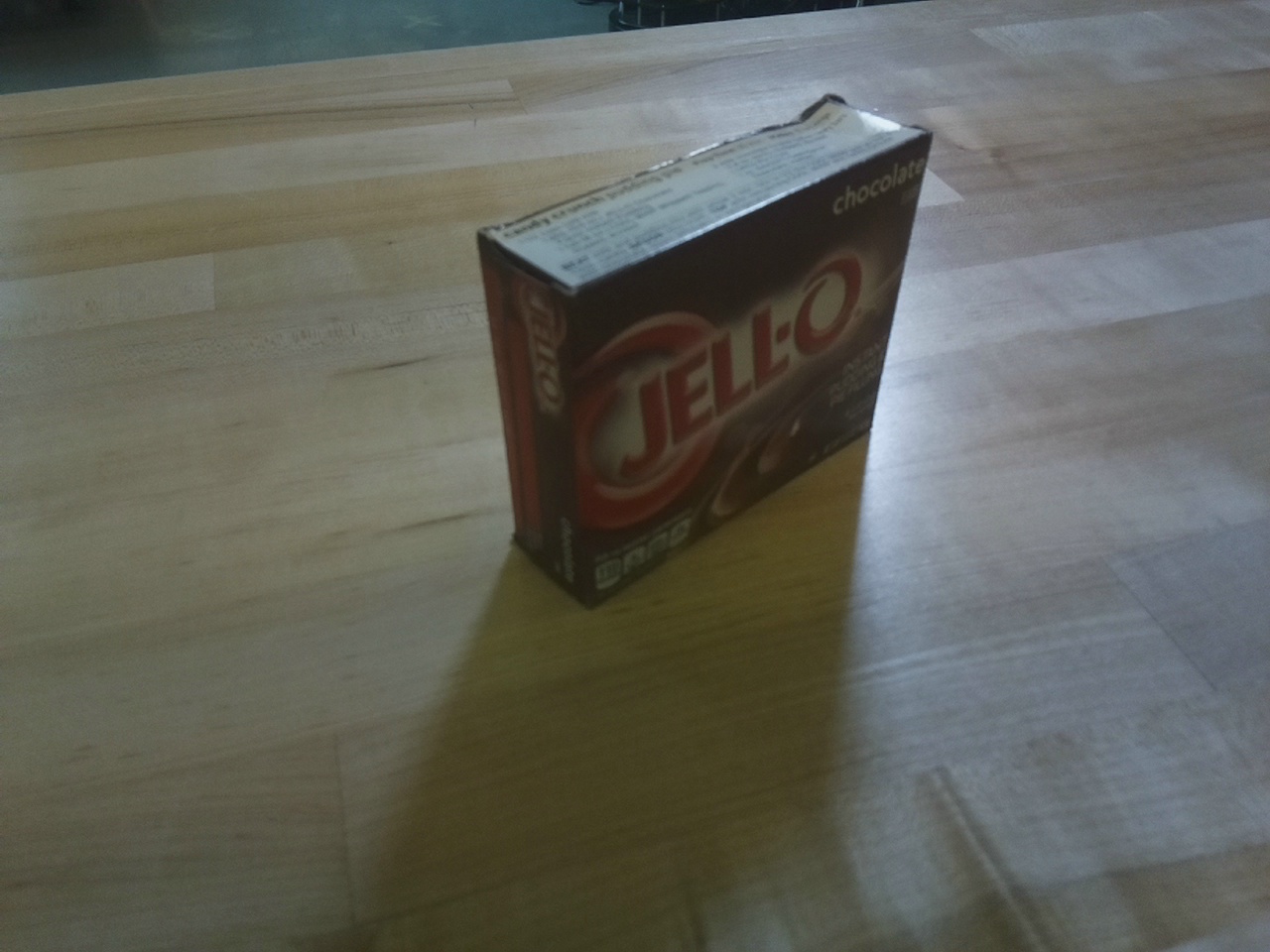}
	\caption{box}
	\end{subfigure}
	\begin{subfigure}[b]{0.11\textwidth}
	\includegraphics[width=\textwidth]{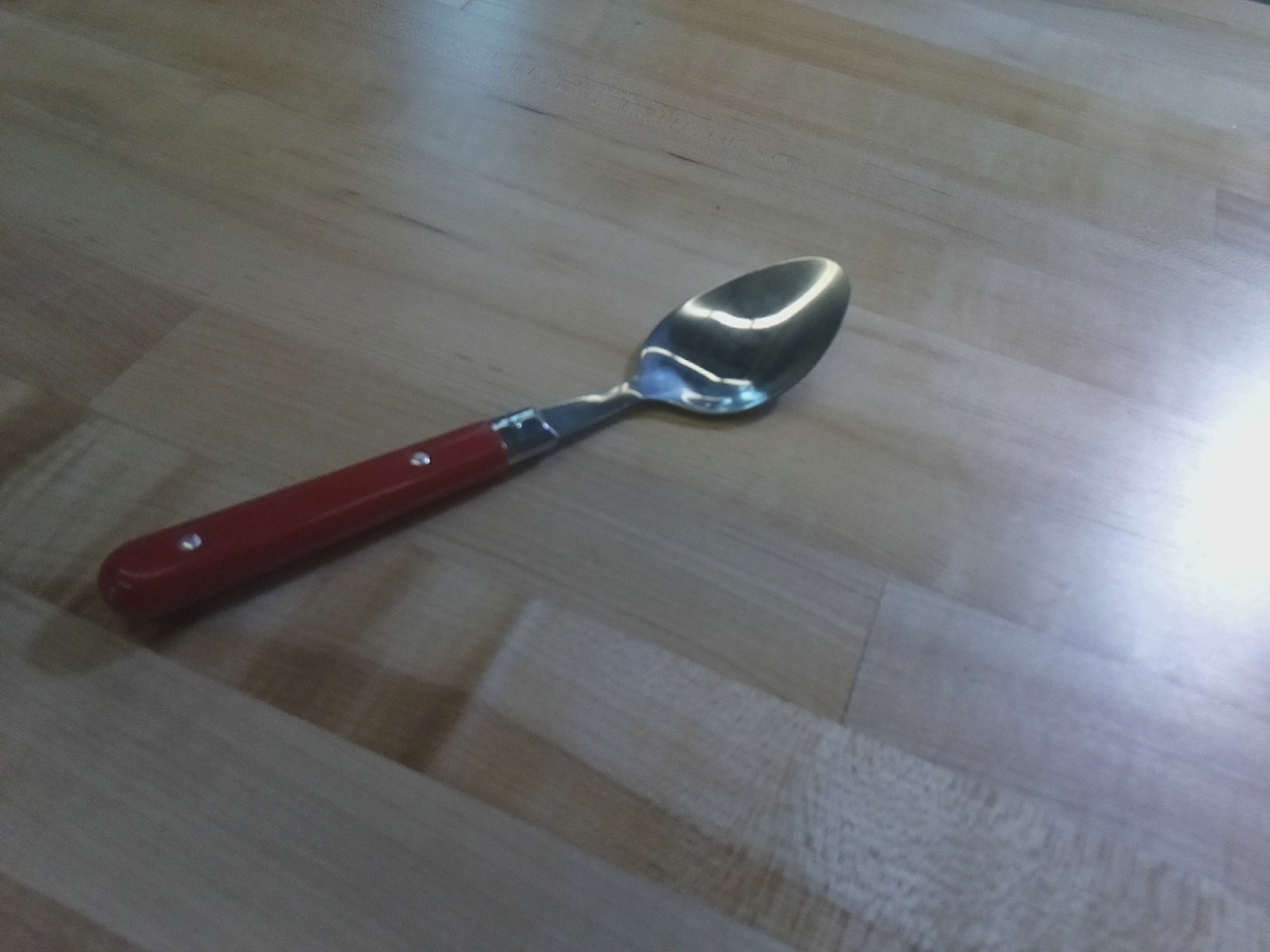}
	\caption{spoon}
	\end{subfigure}
	\begin{subfigure}[b]{0.11\textwidth}
	\includegraphics[width=\textwidth]{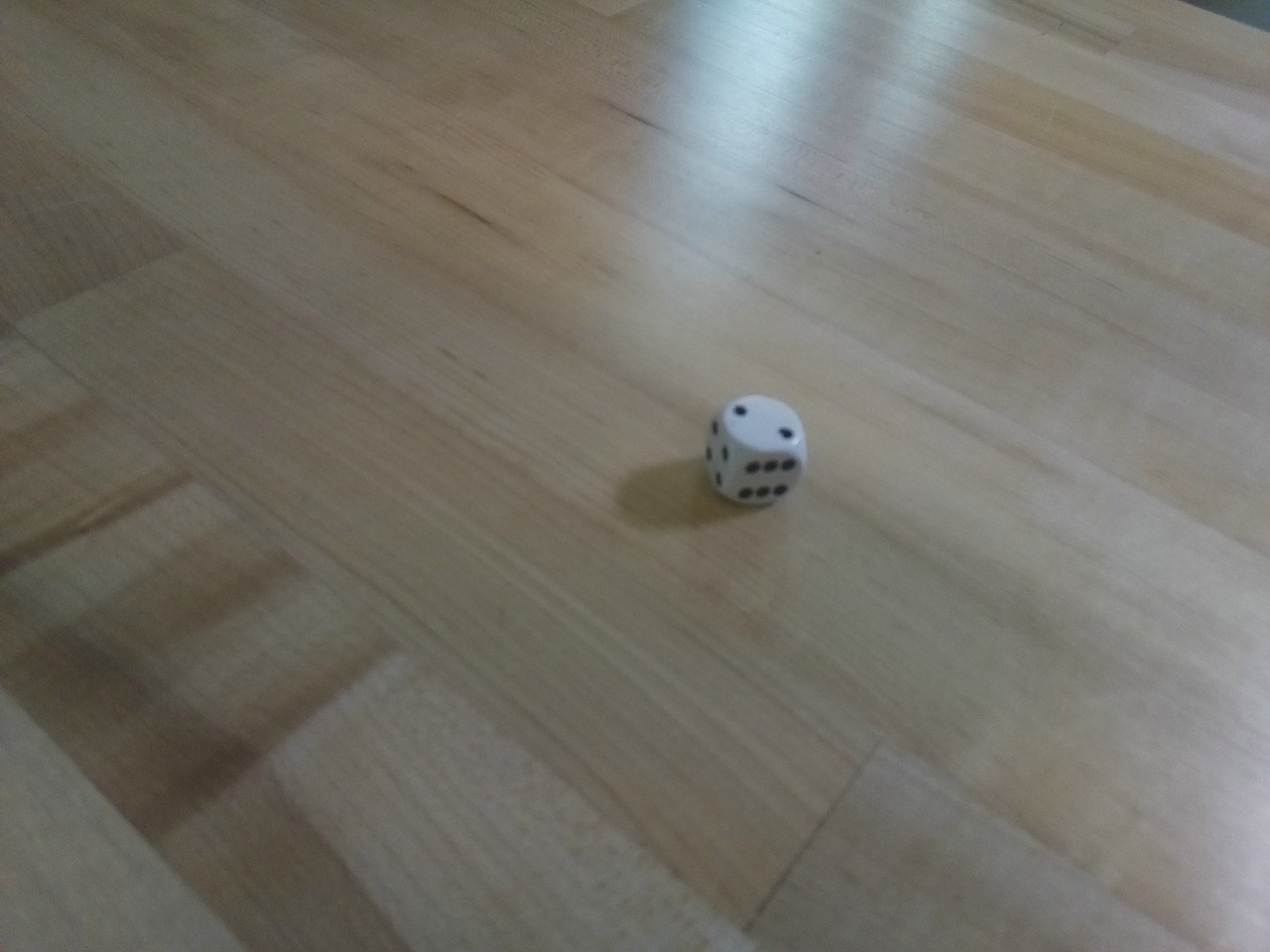}
	\caption{dice}
	\end{subfigure}
	\begin{subfigure}[b]{0.11\textwidth}
	\includegraphics[width=\textwidth]{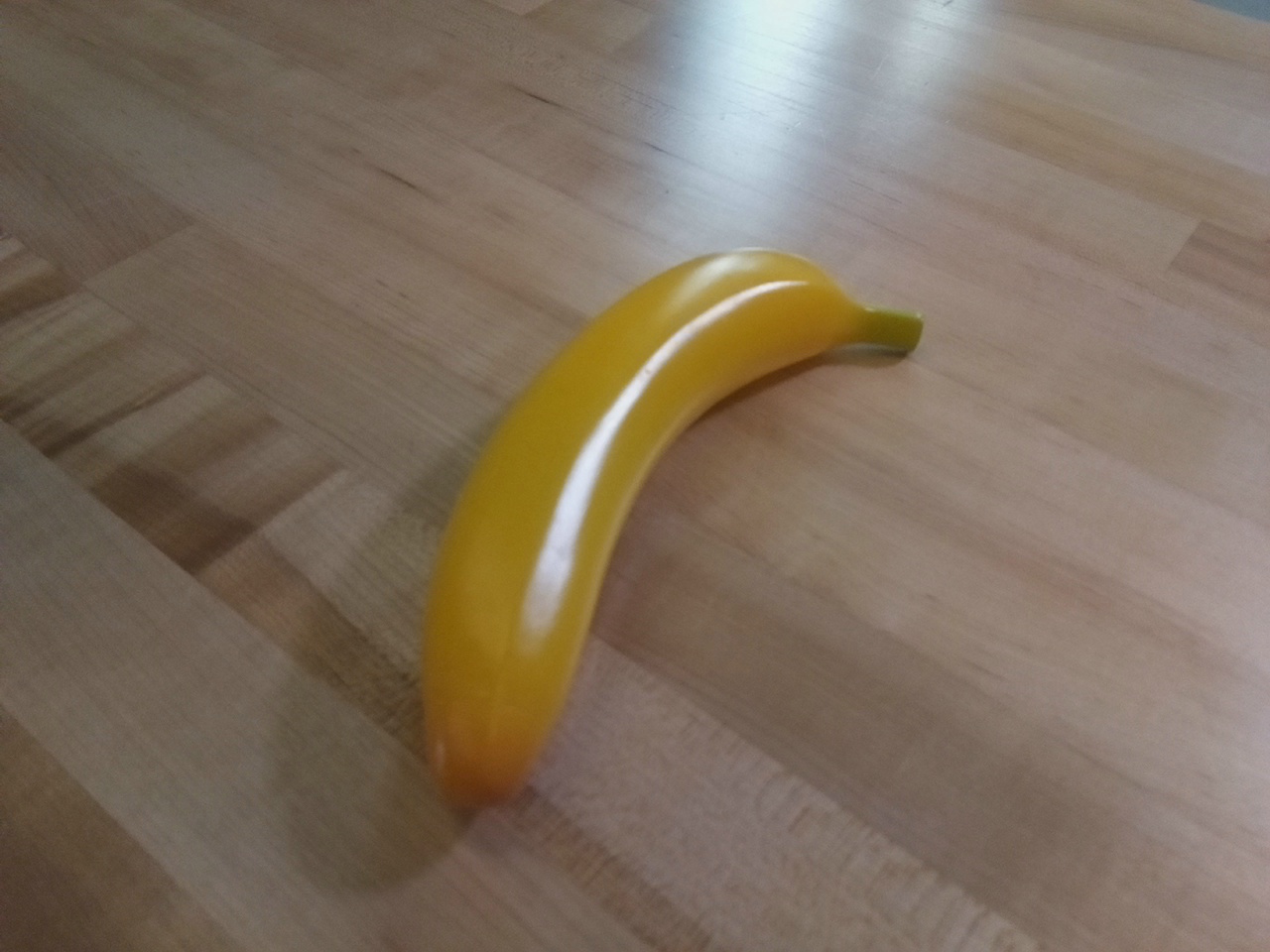}
	\caption{banana}
	\end{subfigure}
}
	\caption{Examples of images collected for training the grasp classifier for the integrated system evaluation.}
	\label{fig:set}
\end{center}
\end{figure}

In the second experiment, the adaptation of the grasping controller was evaluated using seven objects from the YCB dataset \cite{calli2015} and candidate grasps were associated with each. They included an apple (spherical), a cup (cylindrical), a pitcher (hook), a gelatin box (lateral), a spoon (tripod), a dice (pinch), and a banana (tripod). Representative images collected for training are illustrated in Figure \ref{fig:set}. A training set composed of apple, cup, pitcher, box, spoon, and dice images and their corresponding grasps acquired by the adaptive prosthetic hand was used to train the initial neural network. The banana images and their corresponding grasps are not initially used to simulate encountering a novel object. New training examples acquired through the adaptive prosthetic hand are incrementally provided and the inference accuracy of the grasping controller is evaluated to simulate adaptation to a novel object. Inference accuracy of all objects is re-evaluated to test whether the grasp classifier can incorporate the new examples without significantly impacting previous performance. The integrated system evaluation also tests whether the sEMG, audial, and visual information can correctly be classified and execute the correct actions.

\begin{figure}[htb]
\begin{center}
\mbox{
\begin{subfigure}	[b]{0.23\textwidth}
	\includegraphics[width=\textwidth]{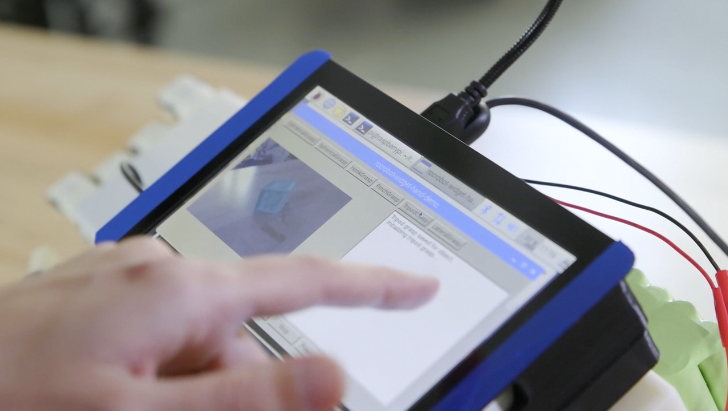}
	\caption{Pressing the ``tripod'' button.}
\end{subfigure}
\begin{subfigure}[b]{0.23\textwidth}
	\includegraphics[width=\textwidth]{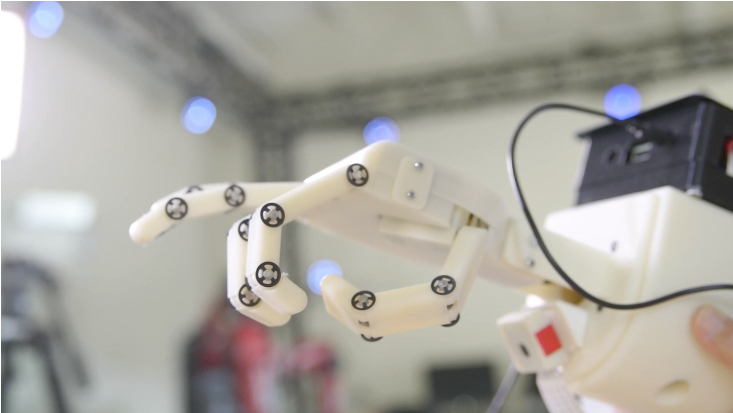}
	\caption{The executed ``tripod'' grasp.}
\end{subfigure}
}
	\caption{Evaluation of the integrated system in the context of the selection of a ``tripod'' grasp through the GUI.}
	\label{fig:guiresults}
\end{center}
\end{figure}

\section{Results}
This section presents the results of the experiments corresponding to the touchscreen and audial interface and adaptation of the integrated system. Qualitative performance of the adaptive prosthetic hand are also discussed. 

For the touchscreen evaluation, each grasp type button was selected through the GUI ten times to verify the performance of that interface. As expected, ten images were saved to the appropriate grasp type. The live feed image of the camera worked appropriately and images corresponding to the correct grasp type were saved correctly with no interruptions during these ten experiments. Feedback was also provided through the touchscreen interface to communicate the behavior of the adaptive prosthetic hand.  Additional options for ``stop'' and ``power down'' were added to the interface to safely stop the motors and return the hand back to the zero, or open hand, position before turning the device off. 

For the language interaction evaluation, a dictionary of words are put together to define the pronunciations of all the words used in each of the sentences. Fourteen annotated examples composed of a variety of different phrases are trained for the language module. These sentences are tested ten times into the microphone. Nineteen untrained sentences are also tested ten times each.  The sentences of the training examples and nineteen novel sentences were provided to the natural language interface.  Success is determined when the uttered commands are correctly translated from speech to text, grounded to the correct grasp, and sent to the controls causing the prosthetic to perform that grasp. A new image should also be saved and labeled to the appropriate grasp type. A failure occurs either due to the speech not being recognized or a failure in parsing the command.  

\begin{table}[ht]
	\begin{center}
	\scriptsize
	\begin{tabular}{c c c c}
		\hline \hline
		Trained Sentences & \%Success & \%Fail-Speech & \%Fail-Parse \\ [0.5ex]
		\hline
		Perform a spherical grasp. & 100 & 0 & 0  \\ 
		Perform a cylindrical grasp. & 90 & 10 & 0  \\
		Perform a hook grasp. & 50 & 50 & 0  \\ 
		Perform a lateral grasp. & 80 & 20 & 0  \\ 
		Perform a tripod grasp. & 100 & 0 & 0  \\ 
		Perform a pinch grasp. & 100 & 0 & 0  \\ 
		Perform a three finger grasp. & 0 & 0 & 100  \\ 
		Perform a two finger grasp. & 0 & 0 & 100 \\ 
		Do a spherical grasp. & 100 & 0 & 0 \\ 
		Open Hand. & 100 & 0 & 0 \\ 
		Close Hand. & 100 & 0 & 0 \\ 
		Open Fingers. & 100 & 0 & 0 \\ 
		Close Fingers. & 100 & 0 & 0 \\ 
		Stop Hand. & 100 & 0 & 0 \\
	\end{tabular}
	\caption{Experiments on interpreting grasping instructions.}
\end{center}
\end{table}

The average runtime of the DCG for symbol grounding in these experiments was 0.011 seconds. Table 1 presents the results for the trained sentences. Three of the sentence failures which occurred were due to speech recognition. If the command was stated too quickly or with hesitation or the words were pronounced differently from the dictionary of words, the speech was not recognized. ``Perform a two finger grasp'' and ``Perform a three finger grasp'' failed due to parsing. One source of this error was due to an incomplete grammar specification that did not specify that specific phrase/word pattern. Later including this corrected the error and allowed for a successful parsing. 

For the untrained sentences, ``Do a cylindrical grasp'' and ``Do a hook grasp'' failed one out of ten times. Speech was not recognized in the failed sentences due to the same reasons described previously. For the recognized speech in these untrained sentences, the novel combinations of observed verbs and noun phrases were able to be correctly inferred as expected.  Novel phrases, such as ``open hand'', ``close hand'', ``make'', and ``give'' incurred parsing errors (when the words or patterns did not appear in the grammar) or understanding errors (when correct parses were generated but training examples were not provided to understanding phrases in the context of their child symbols) as expected.

\begin{figure}[htb]
\begin{center}
	\includegraphics[width=0.48\textwidth]{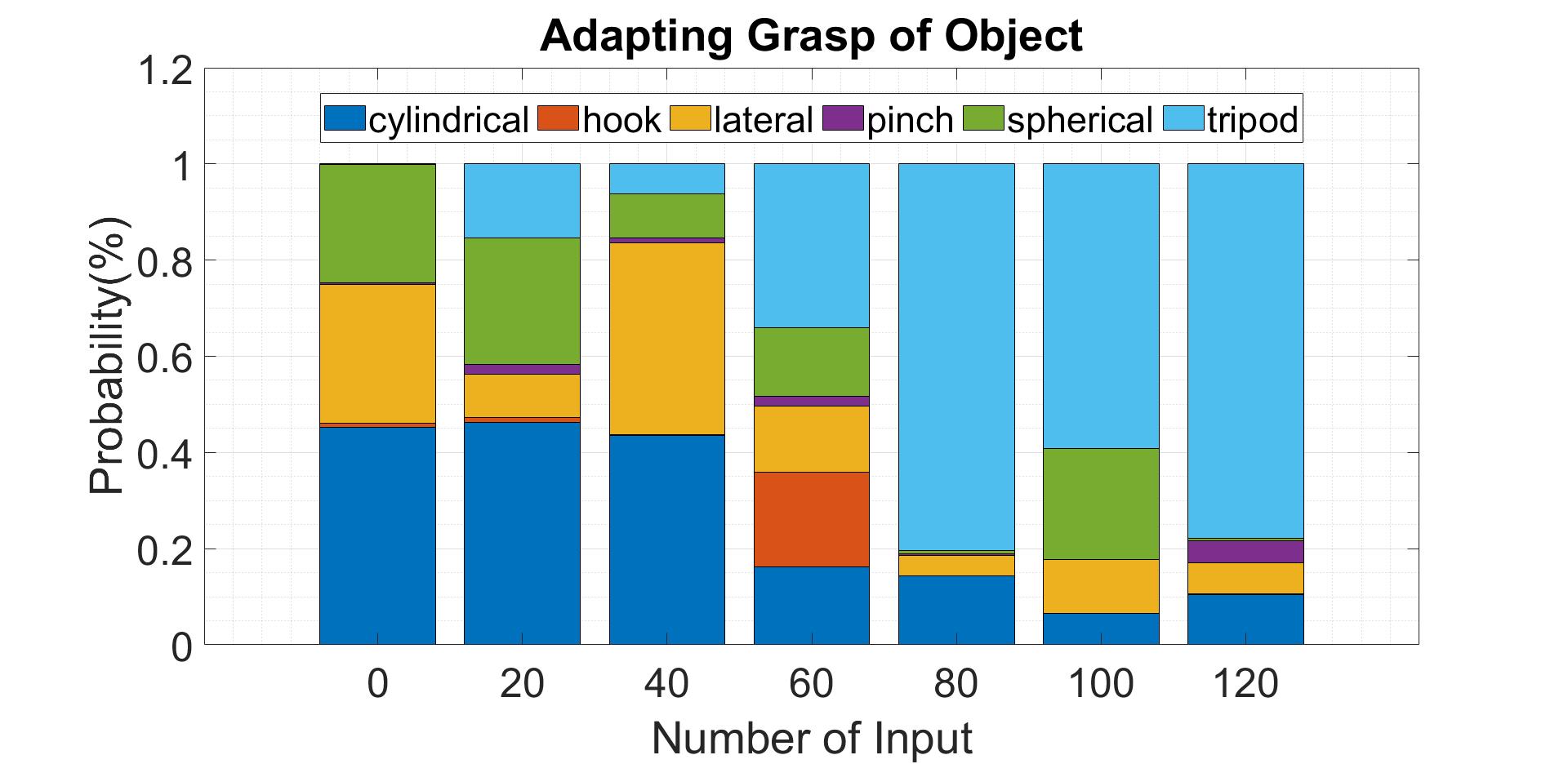}
	\caption{Predicted grasps of the ``banana'' object as a function of data collected in-situ.}
	\label{fig:data4}
	\end{center}
\end{figure}

\begin{figure}[htb]
\begin{center}
\mbox{
\begin{subfigure}	[b]{0.48\textwidth}
	\includegraphics[width=\textwidth]{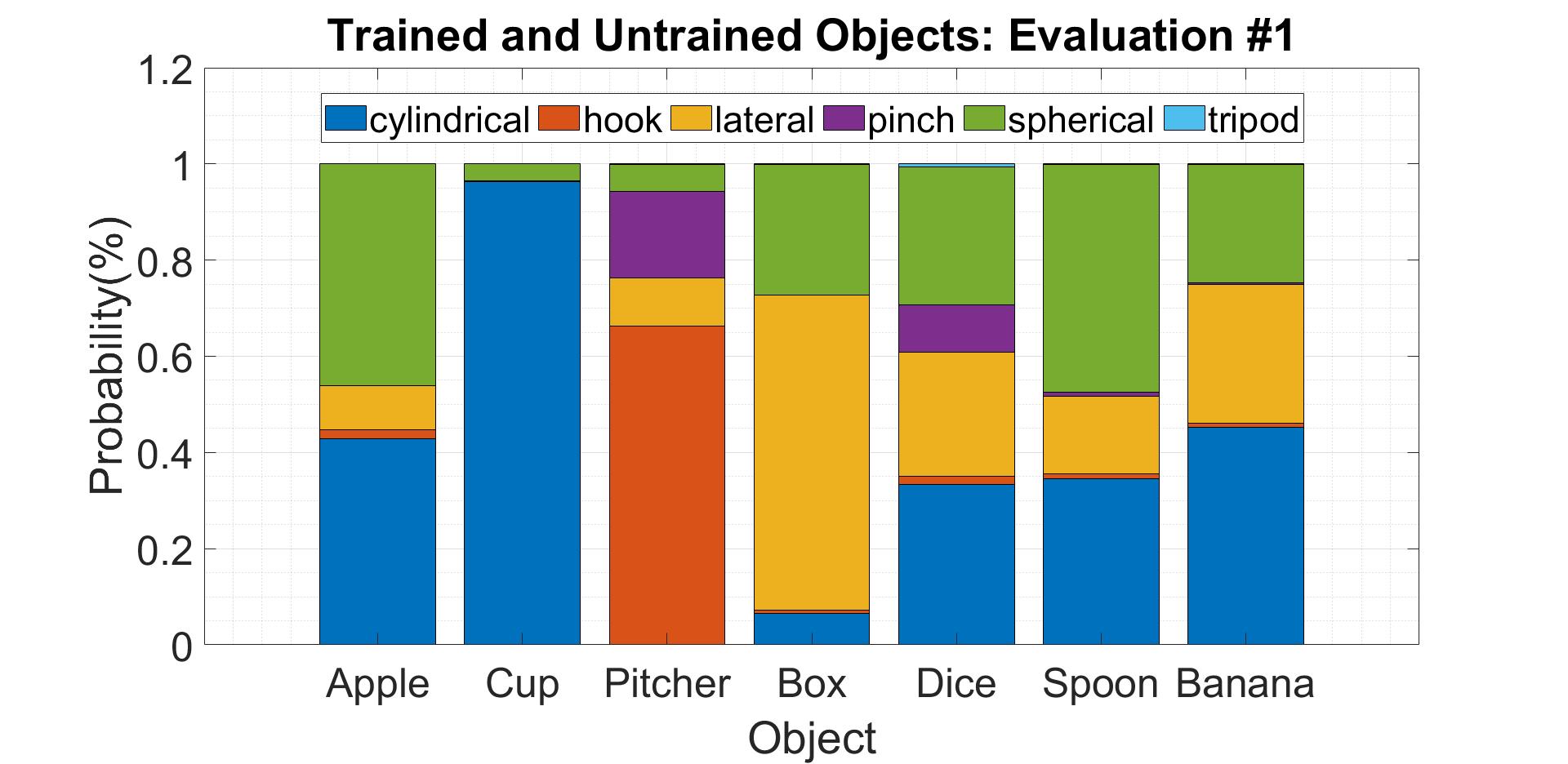}
	\caption{Initial prediction of grasps by the grasping controller.}
	\label{fig:data3}
\end{subfigure}
} \\[6pt]
\mbox{
\begin{subfigure}[b]{0.48\textwidth}
	\includegraphics[width=\textwidth]{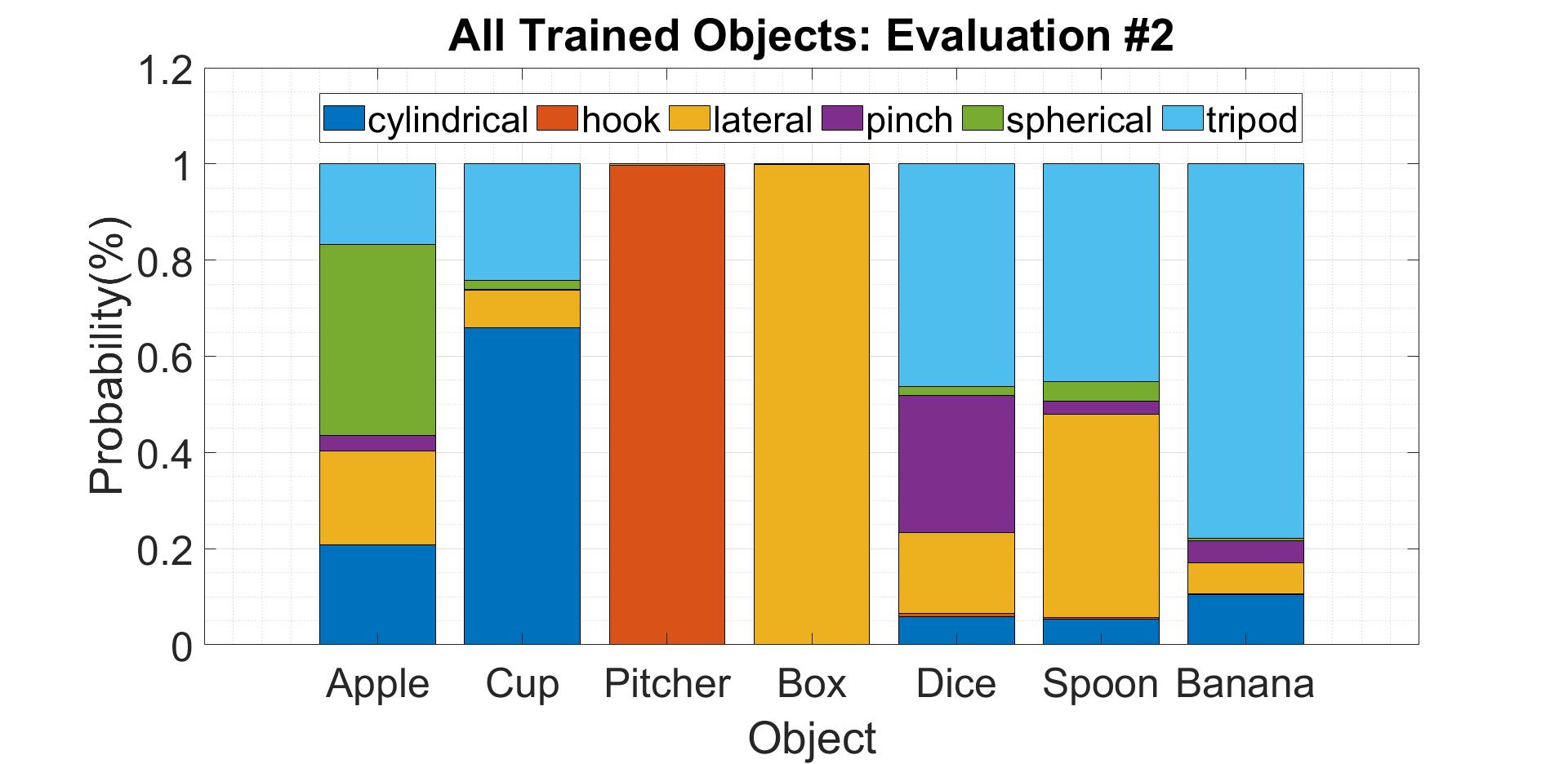}
	\caption{Predicted grasps after addition of examples collected in-situ.}
	\label{fig:data5}
\end{subfigure}
}
	\caption[Adaptation of the grasping controller from examples collected in-situ.]{Adaptation of the grasping controller from examples collected in-situ.}
	\label{fig:adaptation}
	\end{center}
\end{figure}

A total of 120 training examples of images for each of the six out of seven objects were collected using the touchscreen interface. To demonstrate the adaptability of the model, the grasp classifier was trained on these six objects and evaluated on all seven to quantify the performance on an untrained object (the ``banana''). Increments of twenty training examples for this object are added and an evaluation of the grasp classifier performance is performed until one hundred and twenty examples were added. Figure \ref{fig:data3} shows the results after the initial training of six of the seven objects where each was tested ten times. The apple (spherical), cup (cylindrical), pitcher (hook), and gelatin box (lateral) all inferred the correct grasp type as the most probable grasp choice. However, the dice, trained to be classified as a pinch grasp, gave the desired grasp only ten percent of the time. The spoon, trained to be classified as a tripod grasp, was never classified to the desired grasp during testing. This object was classified forty-eight percent of the time as a spherical grasp instead. The banana, the novel object in this experiment, was never classified as a tripod grasp (the desired grasp) using this initial model. Once twenty training examples of the novel object to the corresponding grasp are included and the model is retrained from the augmented dataset, the tripod grasp is predicted with a sixteen percent probability. Retraining is currently performed offline due to the computational limitations of the on-board computer. The grasp classifier performance as a function of additional training examples is illustrated in Figure \ref{fig:data4}. 

\begin{figure}[htb]
	\begin{center}	
\mbox{
    \begin{subfigure}[b]{0.20\textwidth}
	\includegraphics[width=\textwidth]{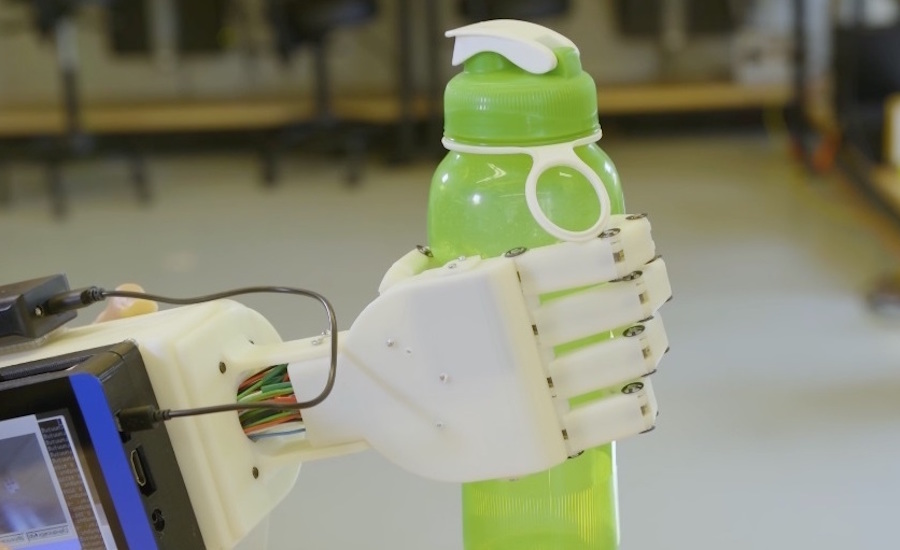}
	\caption{cylindrical}
	\end{subfigure}
	\begin{subfigure}[b]{0.20\textwidth}
	\includegraphics[width=\textwidth]{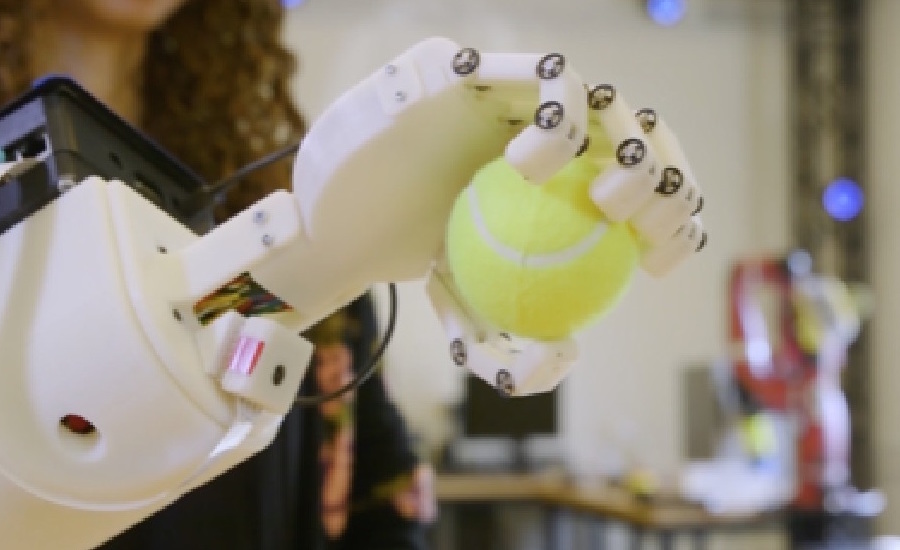}
	\caption{spherical}
	\end{subfigure}
}
\mbox{
	\begin{subfigure}[b]{0.20\textwidth}
	\includegraphics[width=\textwidth]{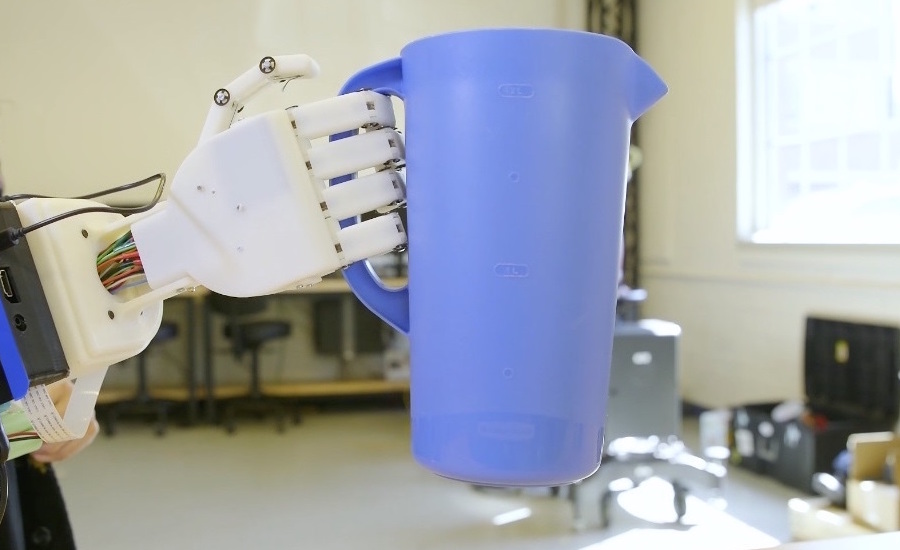}
	\caption{hook}
	\end{subfigure}
	\begin{subfigure}[b]{0.20\textwidth}
	\includegraphics[width=\textwidth]{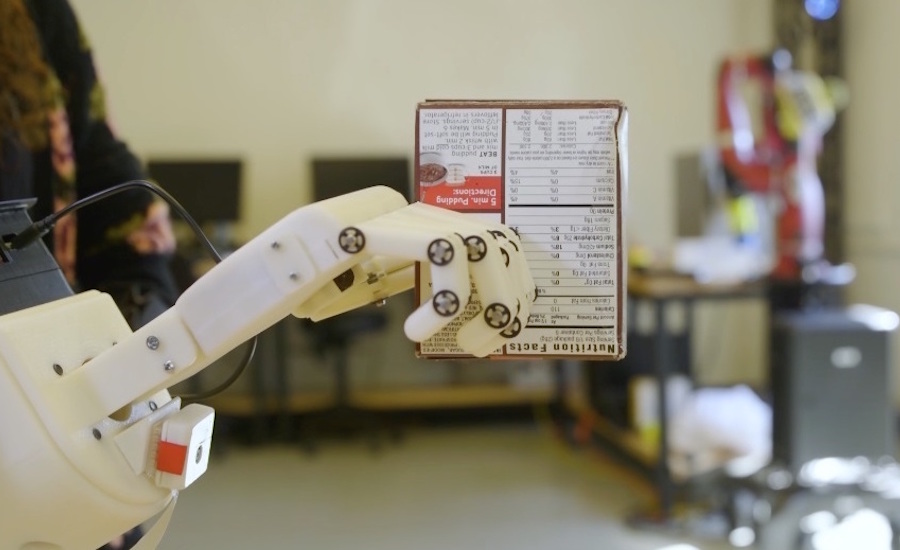}
	\caption{lateral}
	\end{subfigure}
} \\[8pt]
\mbox{
	\begin{subfigure}[b]{0.20\textwidth}
	\includegraphics[width=\textwidth]{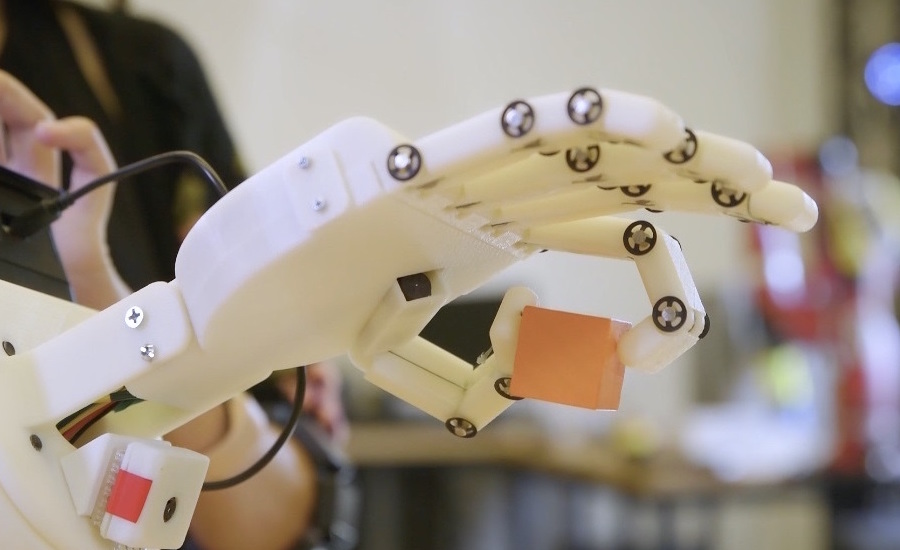}
	\caption{pinch}
	\end{subfigure}
	\begin{subfigure}[b]{0.20\textwidth}
	\includegraphics[width=\textwidth]{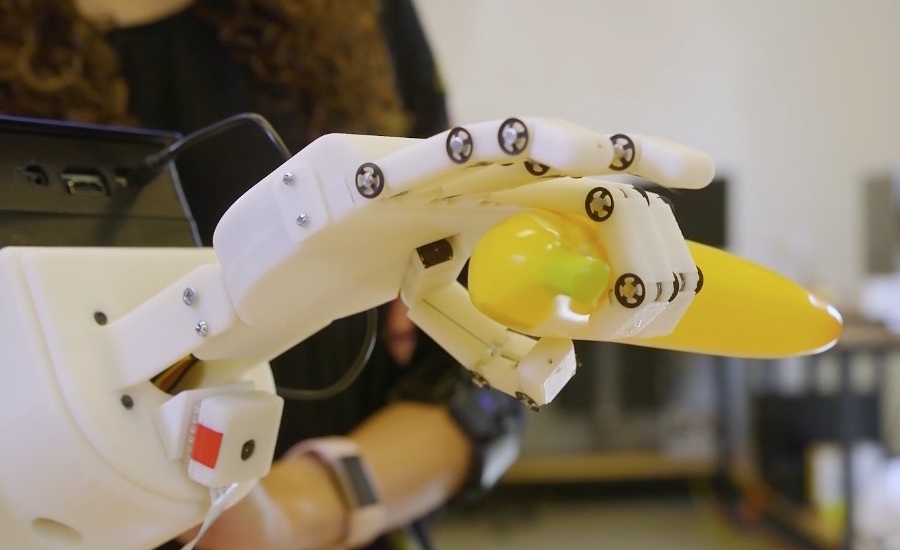}
	\caption{tripod}
	\end{subfigure}
}
	\caption{The adaptive prosthetic hand executing the six grasp types and holding objects from the YCB dataset.}
	\label{fig:graspingresults}
	\end{center}
\end{figure}

After 120 images are added to the neural network and trained, tripod grasp is correctly inferred as most likely for the banana with seventy-eight percent probability. To ensure the grasps of the other objects were not hindered after adding in new data from a different object, all seven of the objects were evaluated. Figure \ref{fig:data5} shows the apple, cup, pitcher, and gelatin box infer correct grasp shapes as the initial model. The desired grasp of the spoon increases in probability from zero percent to forty-five percent. The desired grasp of the dice also increases in probability from ten percent to twenty-nine percent. An explanation for this occurrence could be due to the addition of more data allowing the unique differences in these objects to be more evident. Figure \ref{fig:graspingresults} demonstrates these objects being successfully grasped and held. This experiment demonstrates the main contribution of the approach, which is the framework that enables a prosthetic hand prototype to adapt its behavior through a channel provided by the multi-modal interface in-situ.  

\section{Conclusions}
Many unique control models for prosthetic hands have been developed to address important challenges found by people who use such devices. This paper presents a novel approach to adaptive grasp control which demonstrates the effectiveness of incorporating sEMG signals, visual information, and a multi-modal interface consisting of touchscreen and speech inputs to control and interactively teach a prosthetic device. A prototype of a prosthetic hand device was designed, modeled, fabricated and tested to demonstrate practical application of the interactive learning-based adaptive grasp control. The results presented in this paper show a change in grasping behavior using data acquired through the multi-modal interface.  These experiments support our goal of developing an approach for prosthetic device control that is able to customize its behavior to the preferences of the operator and create a model that is unique to specific users. 

For future work, the use of more advanced perception algorithms would improve the classification performance for more general objects. The integration of force sensors to the finger tips would improve feedback and enable the device to grab objects of varying sizes, weight, and fragility.  Modification of the surface material properties may improve the gripping performance.  More sophisticated natural language models would enable more direct control over grasping behaviors, such as individual finger positions and grasp forces.  Further iterations of the mechanical design would improve the robustness of the experimental platform for longer-duration evaluations of the adaptive grasp controller.  Lastly, incorporating more powerful embedded computers would increase the complexity of algorithms that could be performed and enable on-board model retraining.

\section*{Acknowledgment}
This work was supported by the New York State Center of Excellence in Data Science at the University of Rochester.

\bibliography{esponda_amph_aaai}
\bibliographystyle{aaai}

\end{document}